\documentclass{article}
\usepackage{ijcai25}

\usepackage{times}
\usepackage{soul}
\usepackage{url}
\usepackage[hidelinks]{hyperref}
\usepackage[utf8]{inputenc}
\usepackage[small]{caption}
\usepackage{graphicx}
\usepackage{amsmath}
\usepackage{amsthm}
\usepackage{booktabs}
\usepackage{algorithm}
\usepackage{algorithmic}
\usepackage[switch]{lineno}
\usepackage{multirow}
\usepackage{colortbl}
\usepackage{xcolor}         
\usepackage{amssymb}
\usepackage{mathrsfs}
\usepackage{mathtools}
\usepackage{bm}
\usepackage{tcolorbox}
\usepackage{booktabs}
\newtcolorbox[auto counter]{mybox}[1][]{
  title=My Box \thetcbcounter,
  label=mybox:\thetcbcounter,
  #1
}

\usepackage{newfloat}
\usepackage{listings}
\newcommand{\methodname}{DER}
\newcommand{\eg}{\textit{e.g.}}

\title{Efficient Dynamic Ensembling for Multiple LLM Experts}

\author{
Jinwu Hu$^{1,2*}$
\and
Yufeng Wang$^{1,2,3*}$\and
Shuhai Zhang$^{1,2*}$\and
Kai Zhou$^{1}$\and
Guohao Chen$^{1,2}$\and \\
Yu Hu$^{4}$\and
Bin Xiao$^{5\dagger}$\and
Mingkui Tan$^{1,6\dagger}$
\affiliations
$^1$South China University of Technology\\
$^2$Pazhou Laboratory \\
$^3$Peng Cheng Laboratory\\
$^4$Hong Kong Polytechnic University\\
$^5$Chongqing University of Posts and Telecommunications\\
$^6$Key Laboratory of Big Data and Intelligent Robot, Ministry of Education\\
\emails
fhujinwu@gmail.com, mingkuitan@scut.edu.cn
}

\begin{document}

\maketitle

\begin{abstract}
LLMs have demonstrated impressive performance across various language tasks. However, the strengths of LLMs can vary due to different architectures, model sizes, areas of training data, etc. Therefore, ensemble reasoning for the strengths of different LLM experts is critical to achieving consistent and satisfactory performance on diverse inputs across a wide range of tasks. However, existing LLM ensemble methods are either computationally intensive or incapable of leveraging complementary knowledge among LLM experts for various inputs. In this paper, we propose an efficient \textbf{D}ynamic \textbf{E}nsemble \textbf{R}easoning paradigm, called \textbf{DER} to integrate the strengths of multiple LLM experts conditioned on dynamic inputs. Specifically, we model the LLM ensemble reasoning problem as a Markov Decision Process, wherein an agent sequentially takes inputs to request knowledge from an LLM candidate and passes the output to a subsequent LLM candidate. Moreover, we devise a reward function to train a DER-Agent to dynamically select an optimal answering route given the input questions, aiming to achieve the highest performance with as few computational resources as possible.  Last, to fully transfer the expert knowledge from the prior LLMs, we develop a Knowledge Transfer Prompt that enables the subsequent LLM candidates to transfer complementary knowledge effectively. Experiments demonstrate that our method uses fewer computational resources to achieve better performance compared to state-of-the-art baselines. Code and appendix are available at \href{https://github.com/Fhujinwu/DER}{https://github.com/Fhujinwu/DER}.

\end{abstract}

\section{Introduction}

Large Language Models (LLMs) such as LLaMA \cite{touvron2023llama} and GPT-3.5 \cite{openai2023gpt} have exhibited remarkable performance across diverse tasks \cite{stark2023dobby,hu2025dynamic,wang2025generating}, such as embodied intelligence \cite{mu2023embodiedgpt}. However, their variations in architectures, model sizes, and training data result in distinct strengths and weaknesses in different tasks \cite{jiang2023llm,lu2023routing}. Although training a larger LLM with more comprehensive data is possible to maintain excellent performance in all tasks, the cost is significant and often impractical. Consequently, it is crucial to assemble LLMs to enhance their generalization while minimizing the consumption of computational resources in practical applications.

\textit{Unfortunately}, assembling knowledge of LLMs with limited computing cost is difficult partly for the following reasons. 1) \textit{Complex knowledge integration}: Each LLM is typically trained on different datasets, which may include varying levels of quality, diversity, and bias. Harmonizing these LLMs requires aligning their understanding with knowledge bases, posing challenges to integration \cite{jiang2023llm,lu2023routing}. 2) \textit{High computational complexity}: LLMs are computationally intensive, requiring significant resources for inference. Combining LLMs increases this complexity, potentially necessitating more efficient algorithms to manage \cite{sheng2023flexgen,wan2024knowledge}.

\begin{figure*}[t]
\centering
\includegraphics[width=0.8034\linewidth]{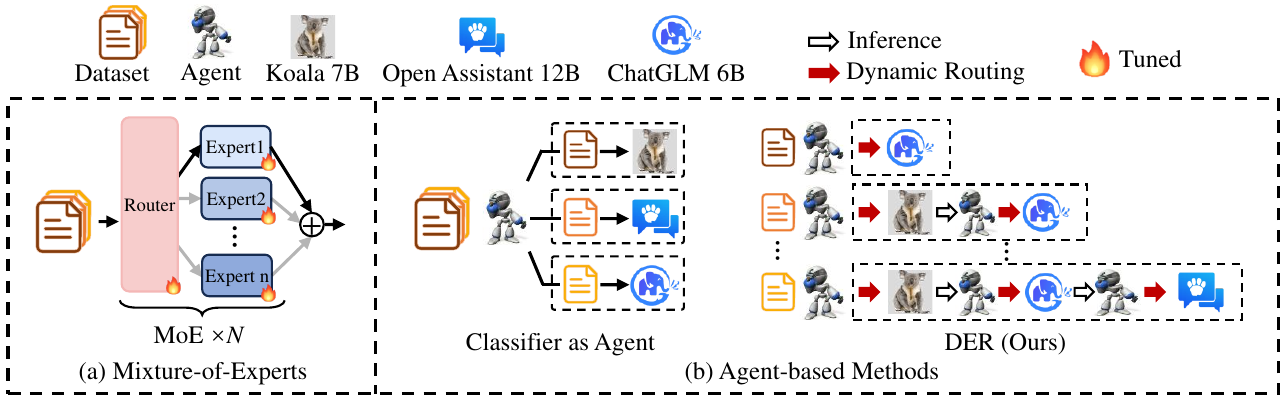}
    \caption{
    Illustration of different LLM ensemble strategies.
    \textbf{(a)} Ensemble with MoEs. \textbf{(b)} Ensemble with the agent.
    }
    \label{fig: intuition and motivation}
\end{figure*}

Recently, LLM ensemble has emerged as a rapidly progressing research area to leverage the diverse strengths of multiple LLMs. Based on their strategies, most existing methods can be broadly divided into four types. 
The \textit{Mixture-of-Experts} methods \cite{jiang2024mixtral,tang2024merging} use a router network to select and activate a subset of experts to aggregate diverse expertise (see Figure \ref{fig: intuition and motivation} (a)), but they often require retraining, cannot integrate non-homologous LLM experts, and fail to leverage complementary knowledge among experts.
The \textit{parameter merging} methods~\cite{yu2023language,matena2022merging} merge the parameters of homologous LLMs into a single unified model but cannot assemble non-homologous LLMs. The \textit{rule-based} methods~\cite{dong2023self,du2023improving} assemble the advantages of LLM by manually designing task-specific roles or a fixed order. However, such a static setting makes it difficult for the integration to generalize in various domains. To avoid these issues, \textit{agent-based} methods~\cite{jiang2023llm} train an agent to integrate non-homologous LLMs with the various strengths, making them adaptable to various scenarios.

Despite the recent success of agent-based LLM ensemble methods in outperforming the best one among the LLMs across a wide range of downstream tasks, these methods still face certain limitations. Firstly, a common drawback in most existing methods~\cite{jiang2023llm,liu2023dynamic} is that they integrate LLMs based on the final outputs of all LLMs. These methods require prohibitively high computational costs to run all the LLM candidates and result in inefficient utilization of inference resources. Therefore, some easy samples that could be simply addressed by a single LLM have to run all the LLMs at expensive costs. For example, PairRanker~\cite{jiang2023llm} requires significantly more parameters ($>117B$) than a single LLM in inference, potentially leading to unbearable resource wastage. Secondly, while some agent-based methods implement a classifier to accomplish the integration by selecting only one LLM candidate at each time with very little inference cost (see Figure \ref{fig: intuition and motivation}(b), left), \eg, ZOOTER \cite{lu2023routing}, their performance is limited by the fact that they do not leverage the complementary knowledge among LLMs.

To address the above limitations, we propose a novel \textbf{D}ynamic \textbf{E}nsemble \textbf{R}easoning paradigm for integrating the strengths of multiple LLM experts, called \textbf{DER}. Given that LLMs are always trained on diverse datasets, we hypothesize that they possess complementary knowledge, which can be sequentially assembled. Nevertheless, the exponential growth in possible combinations of routes renders it impractical to address this challenge through classification tasks alone. To overcome this, we view knowledge transfer as a sequential decision process. Specifically, we model the LLM ensemble as a Markov Decision Process (MDP), where a DER-Agent dynamically requests contributions from LLMs and transfers this knowledge to subsequent LLM candidates (see Figure \ref{fig: intuition and motivation}(b), right). Moreover, we develop a reward function to train the DER-Agent to select optimal answering routes based on input questions, aiming to maximize performance while minimizing computational resources. Additionally, we introduce a Knowledge Transfer Prompt (KTP) to facilitate effective knowledge transfer among LLMs.

We summarize our main contributions as follows:
\begin{itemize}
\item
We propose Dynamic Ensemble Reasoning (DER) for the LLM ensemble, modeling it as a Markov Decision Process (MDP) for efficient sequential knowledge transfer. This approach dynamically selects optimal answering routes, integrating complementary knowledge from various LLMs to maximize performance with minimal computational resources. Experiments show DER integrates the strengths of different LLMs, achieving nearly a \textit{7-fold} parameter reduction compared to ensemble methods using the outputs of all LLMs.
\item 
We introduce a Knowledge Transfer Prompt (KTP) to facilitate efficient knowledge sharing among LLMs and develop a reward function to train the DER-Agent. This ensures the DER-Agent leverages expert knowledge from previous LLMs, optimizing task performance while significantly reducing computational costs. Experiments show that more than 9\% improvement is achieved on BARTScore using KTP and our reward function.
\end{itemize}

\begin{figure*}[t]
\centering
\includegraphics[width=0.77\linewidth]{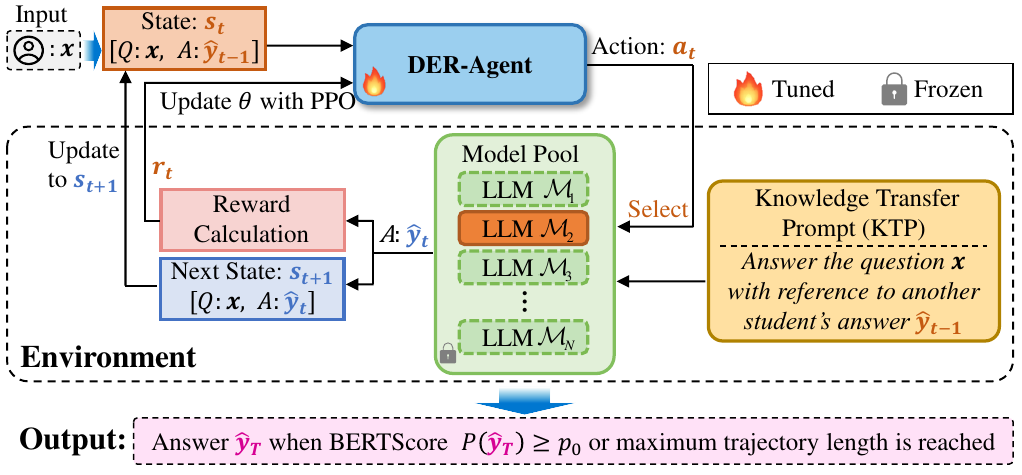}
    \caption{
    General diagram of DER. We formulate the LLM ensemble as an MDP and train DER-Agent to select an optimal answering route for inputs. At step $t$, the DER-Agent takes ${\color[RGB]{197, 90, 17}s_t}=[Q: x, A: {\color[RGB]{197, 90, 17}\hat{y}_{t-1}}]$ as input and selects an LLM $\color[RGB]{197, 90, 17}\mathcal{M}_{a_t}$ to continue answering the question regarding the existing answer, leading to a new answer $\color[RGB]{68,114,196} \hat{y}_{t}$. We calculate a reward ${\color[RGB]{197, 90, 17}r_t}$ with $\color[RGB]{68,114,196} \hat{y}_{t}$ and update the state to $\color[RGB]{68,114,196}s_{t+1}$. This process will loop until the answer is evaluated as satisfactory enough or the max trajectory length.}  
    \label{fig:mdp}
\end{figure*}

\section{Related Works} 
\textbf{Mixture-of-Experts Methods} integrate the knowledge of LLMs by selecting and activating a subset of experts through a router network. Jiang \textit{et al.}
\cite{jiang2024mixtral} propose a Sparse Mixture of Experts language model, which aggregates the knowledge of diverse experts by selecting two experts at each layer of the routing network to process the current state and combining their outputs. Tang \textit{et al.} \cite{tang2024merging} propose the Weight-Ensemble MoE, which achieves assembling different expert knowledge by training a router to select and merge different LLM expert parameters. However, these methods often require re-fine-tuning of the MoE model, and it is difficult for experts to utilize the complementary knowledge of other experts.

\textbf{Parameter Merging Methods} assemble the advantages of LLMs by merging the parameters of multiple homologous LLMs into a single model. Matena and
Raffel \cite{matena2022merging} propose the ``Fisher merging'', which merges the parameters of models with the same structure and initialization, achieving the ability to assemble different LLMs. Yu \textit{et al.} \cite{yu2023language} introduce an operation called DARE to sparsify delta parameters of multiple homologous LLMs, and subsequently merge them into a single model by parameter averaging, realizing the ability to assemble LLMs. However, these methods cannot assemble non-homologous LLMs. 

\textbf{Rule-based Methods} assemble LLMs by manually setting roles or fixed orders for specific tasks. Du \textit{et al.} \cite{du2023improving} propose LLM-debate, in which multiple LLMs get a consensus in multiple rounds of common debate. Dong \textit{et al.} \cite{dong2023self} use three LLMs, set up as analysts, coders, and testers, to collaboratively develop software in sequential execution. Despite the advantages of these to quickly assemble LLMs, this static setup is hard to generalize.

\textbf{Agent-based Methods} dynamically integrate the advantages of different non-homologous LLMs by training an agent and can be used in various scenarios. Jiang \textit{et al.} \cite{jiang2023llm} propose the LLMs ensemble framework LLM-BLENDER in expectation of consistently superior performance, which selects the TOP-K responses by PairRanker and mixes them to generate final outputs using the GENFUSER. Liu \textit{et al.} \cite{liu2023dynamic} propose DyLAN, which enables the LLM ensemble through multiple rounds of interaction and an early stopping mechanism. However, they are assembled with the outputs of all candidate LLMs, which tends to waste significant computational resources. Lu \textit{et al.} \cite{lu2023routing} propose ZOOTER, a reward-guided routing method that can directly and accurately assign each prompt to LLMs with expertise, using a small computational overhead. However, it is limited in answering performance because it does not utilize complementary knowledge between LLMs.

\section{Proposed Methods}
\subsection{Problem Definition and Motivations}
\textbf{Problem Definition.} Given an input question set $\left\{{x}_1, \ldots, {x}_K\right\}$ with $K$ questions, the LLM ensemble aims to aggregate the strengths of different LLMs (experts) to consistently achieve superior performance. Specifically, given an input question $x$ and a model pool $\left\{\mathcal{M}_1, \ldots, \mathcal{M}_N\right\}$ with $N$ LLMs, we aim to select $k$ models to answer the question together to improve the quality of the answer.

\textbf{Motivations.}
Existing methods \cite{lu2023routing,yao2023solving} train an agent to select the most suitable LLM to answer each question individually (see Figure \ref{fig: intuition and motivation} (b), left). Although these methods can improve the answering performance, their performance is limited due to the underutilization of the different knowledge contained in LLMs. To overcome this issue, we aim to develop an effective knowledge aggregation strategy to achieve superior performance. Intuitively, leveraging knowledge transfer from one LLM to another can compensate for individual LLM shortcomings, fostering a collaborative integration of diverse knowledge. Moreover, there exists an optimal knowledge transfer route for each question. \textit{Unfortunately}, the number of possible route combinations is as large as $\sum_{k=1}^{m} N^k$, where $N$ is the number of LLMs, $k$ is the route length, and $m$ is the maximum route length. In this sense, finding an optimal route for each sample poses a severe challenge.

To address the above issue, we naturally model the knowledge transfer as a sequential decision process, given its sequential nature. To this end, we require selecting certain models consecutively from the pool to answer the question, with each model able to refer to the responses provided by its predecessors (see Figure \ref{fig: intuition and motivation} (b), right). In this way, we derive an ultimate response from the terminal model. Our goal lies in identifying an optimal pathway of model execution $\mathrm{Route}^* {:=} [\mathcal{M}_i {\Rightarrow} \mathcal{M}_j {\Rightarrow} \cdots {\Rightarrow} \mathcal{M}_k]$ to enhance the quality of the final answer with modest computational cost. 
As shown in Figure \ref{fig:mdp}, we formulate the LLM ensemble as a Markov Decision Process (MDP) and train a DER-Agent to select an optimal answering route. At each time step $t$, the DER-Agent determines the next LLM to generate a response based on the input question and the current answer (initially absent). This procedure utilizes the Knowledge Transfer Prompt (KTP), facilitating the LLM to construct an answer that progressively integrates insights from the previous  LLM. The newly formed answer serves as the next state, enabling the ongoing knowledge transfer. This process continues looping until the answer is satisfactory or the maximum trajectory length is reached.

\subsection{Dynamic Ensemble as an MDP}
We seek a general DER-Agent to select an optimal sequential execution route of LLMs for the input question $x$, obtaining the highest performance with the least possible computational resources. 
To this end, we formulate the selection of the sequential execution route of LLMs as Markov Decision Process~(MDP) \cite{van2012reinforcement}: $<\mathcal{S}, \mathcal{A},\bm{\mathcal{T}},\bm{\mathcal{R}},\pi>$. 
The state space of the environment is $\mathcal{S}$ and the action space of the agent is $\mathcal{A}$. At time step $t$, the agent takes the state $s_t\in \mathcal{S}$ as input and performs an action $a_t\in \mathcal{A}$ through the policy network $\pi: \mathcal{S}\times \mathcal{A} \rightarrow \left[0,1\right]$. The environment changes to the next state $s_{t+1}=\bm{\mathcal{T}}(s_{t}, a_{t})$ according to the transition function $\bm{\mathcal{T}}$ and a reward $r_t=\bm{\mathcal{R}}(s_{t},a_{t})$ is received with reward function $\bm{\mathcal{R}}$. The MDP is detailed as:

\textbf{States} $\mathcal{S}$ is a set of states which describe the environment. At time step $t$, the state consists of a question-answer pair: $s_t=[Q: x, A: \hat{y}_{t-1}]$, where $x$ is the input question and  $\hat{y}_{t-1}$ is the answer from the selected LLM at time step $t-1$ (the answer is $None$ initially). Thus, the agent judges the quality of the current answer by the question-answer pair and predicts which next LLM to answer.

\textbf{Actions} $\mathcal{A}$ is a discrete set of actions the agent takes. The action space $\mathcal{A}=\left\{1,2,\dots, N\right\}$ is the index of each LLM. At time step $t$, the agent gives the action $a_t \in \mathcal{A}$ based on the state $s_t$ to select a next LLM $\mathcal{M}_{a_t}$ from the model pool $\{\mathcal{M}_1, \dots, \mathcal{M}_N\}$. 

\textbf{Transition} $\bm{\mathcal{T}}(\mathcal{S},\mathcal{A})$ is a function $\bm{\mathcal{T}}$: $\mathcal{S} \times \mathcal{A} \rightarrow \mathcal{S}$ which maps a state $s_t$ into a new state $s_{t+1}$. When the BERTScore $P(\hat{y}_T)$ for the answer $\hat{y}_T$ reaches a manually-set threshold $p_0$ or the maximum trajectory length $T_{max}$ is reached, this episode will be terminated and $s_{T+1}$ is $None$. Otherwise, the selected LLM at time step $t$ will answer the question $x$ concerning the existing answer $\hat{y}_{t-1}$:
\begin{align}
 s_{t+1}&=[Q: x, A: \hat{y}_{t}], \\
 \mathrm{where}~~\hat{y}_{t}&= \mathcal{M}_{a_t}({KTP}(x,\hat{y}_{t-1})).\notag
 \end{align}
$KTP(\cdot)$ is the Knowledge Transfer Prompt (KTP) that we designed to stitch together the question and answer from the previous LLM $\mathcal{M}_{a_{t-1}}$ and to promote the current LLM to answer the question $x$ concerning the previous answer $\hat{y}_{t-1}$. The KTP is detailed in subsection \ref{subsec: ktp}.

\textbf{Rewards} $\bm{\mathcal{R}}(\mathcal{S},\mathcal{A})$ is the reward function. In the LLM ensemble task, the reward can be considered as an evaluation of the quality of the answer ${\hat y}$ for the selected LLM $M_{a_t}$. The details of the reward function are given in the subsection \ref{subsec: reward}. 

\textbf{Policy} $ \pi_\theta(a\mid s): \mathcal{A} \times \mathcal{S} \rightarrow[0,1]$ describes the behaviors of the agent. During the training process, the agent takes the current state $s_t$ as input and outputs a probability distribution for each possible action $a_t \in \mathcal{A}=\left\{ 1,2,\dots, N\right\}$:
\begin{equation}
    \pi\left(a_t=i \mid s_t ; \theta\right)=\frac{\exp \left\{ f_{\theta}\left(s_t\right)_{i} \right\}}{\sum_{j=1}^N \exp \left\{ f_{\theta}\left(s_t\right)_{j} \right\}},
\end{equation}
where $f_{\theta}\left(s_t\right)$ is the output vector of the policy network with input $s_t$, and $i$ denotes the index of the action. $\theta$ is the learnable parameters of the policy network.

The general diagram of the proposed \methodname~is shown in Figure \ref{fig:mdp}.
Given an input question $x$, the DER is initialized with state $ s_0=[Q: x, A: None]$. The agent takes $s_0$ as input and gives an action $a_0$ so that an LLM $\mathcal{M}_{a_0}$ is selected to answer the question with $\hat{y}_0$. And then the reward $r_0$ is calculated for agent optimization based on the answer's quality and computational resources.
The state is updated to $s_1=[Q: x, A: \hat{y}_0]$ with the answer $\hat{y}_0$. The above process will continue until the BERTScore $P({{\hat y}_t})$ exceeds the threshold $p_0$ or the maximum trajectory length is reached. Finally, the last answer $\hat{y}_T$ is obtained, which is high-quality thanks to the knowledge transfer among LLMs.

\subsection{Reward Function Design}
\label{subsec: reward}
In our designed MDP, the reward function is defined to reflect three aspects: the quality of the answer provided by the selected LLM, the increment quality of the answer, and the computational resources:
\begin{equation}
\label{equation: rt}
\bm{\mathcal{R}}_t = \left\{ {\begin{array}{*{20}{l}}
{P({{\hat y}_t}) - \alpha C(M_{a_t}),\quad \quad  \quad \quad \quad \quad t = 0 }\\
{P({{\hat y}_t})  + \beta \Delta P({\hat y}) - \alpha C(M_{a_t}),   \quad \   t > 0}
\end{array}} \right.,
\end{equation}
where $P{(\cdot)}$ is the BERTScore, which is commonly used to evaluate the quality of generated text and its high correlation with human judgment \cite{zhang2019bertscore}. The $\hat{y}_t$ is the output answer of the selected LLM $\mathcal{M}_{a_t}$, $C(\cdot)$ is the computation cost of $\mathcal{M}_{a_t}$, $\Delta P({\hat y})=P({{\hat y}_t}) - P({{\hat y}_{t - 1}})$ is the increment of the BERTScore of the answer from $t-1$ to $t$, and $\alpha, \beta$ is the coefficient to determine the ratio of computation cost and the increment of the score, respectively. In addition, we add additional rewards or penalties to allow the agent to complete the generation of routes in limited steps. Thus, the complete reward follows:
\begin{equation}
\label{equation: rsa}
\bm{\mathcal{R}}({s_t},{a_t}) = \left\{ {\begin{array}{*{20}{l}}
{\bm{{\mathcal{R}}_t} + \gamma,  \quad t \le T_{max} \  and \  P({{\hat y}_t}) \ge p_0}\\
{{\bm{\mathcal{R}}_t} - \gamma, \quad t = T_{max} \ and \ P({{\hat y}_t}) < p_0}
\end{array}} \right.,
\end{equation}
where $p_0$ is the threshold of the BERTScore for which an environment gives an end, $T_{max}$ is the maximum step size, and $\gamma$ is the bias for extra rewards or penalties. Note that in the testing phase $P({{\hat y}_t}) \ge p_0$ or $P({{\hat y}_t}) < p_0$ is judged by one of our trained Terminator, which is a binary classifier.

\begin{table*}[t]
\renewcommand{\arraystretch}{0.98}
\renewcommand{\tabcolsep}{0.20pt}
\begin{center}
\begin{small}
\begin{tabular}{ccccccccccc}
\toprule[1pt]
\multirow{2}{*}{Category}       & \multicolumn{2}{c}{\multirow{2}{*}{Methods}}    & \multicolumn{2}{c}{Param. $\downarrow$} & \multirow{2}{*}{s/sample $\downarrow$} & \multirow{2}{*}{Rouge-S $\uparrow$} & \multirow{2}{*}{BARTScore $\uparrow$} & \multirow{2}{*}{BLEURT $\uparrow$} & \multirow{2}{*}{BERTScore $\uparrow$} &\multirow{2}{*}{GPT-Rank$\downarrow$}\\ \cline{4-5}
& \multicolumn{2}{c}{}                            & Agent    & Infer.      &              &              &              &              &                       &  \\ \hline
& \multicolumn{2}{c}{Oracle (BARTScore)}    & --       & --            &--           & 0.33  &-2.87      &-0.38  &73.23                 &-  \\
& \multicolumn{2}{c}{ChatGPT \cite{openai2023gpt}}    & --       & $\approx$175B            & --           & 0.39  &-3.00     &-0.26  &76.23                 &-  \\ \hline
\multirow{11}{*}{LLMs}        & \multicolumn{2}{c}{MOSS \cite{sun2023moss}}                        & --       & 16B            & --                      & 0.19   &-3.69                   & -0.73     &64.85             &-  \\
                              & \multicolumn{2}{c}{Vicuna \cite{chiang2023vicuna}}                      & --       & 13B            & --                      & 0.27  &-3.44                    & -0.61                 & 69.60 &-   \\
                              & \multicolumn{2}{c}{Alpaca \cite{taori2023stanford}}                      & --       & 13B            & --                     & 0.29  &-3.57                   & -0.53                  & 71.46 &-  \\
                              & \multicolumn{2}{c}{Baize \cite{xu2023baize}}                       & --       & 13B            & --                      & 0.20   &-3.53                   & -0.66            & 65.57      &-  \\
                            & \multicolumn{2}{c}{Open Assistant \cite{openassistant}}              & --       & 12B            & --                      & 0.34      &-3.45               & -0.39     & 74.68     &-           \\
                            & \multicolumn{2}{c}{Dolly2 \cite{conover2023free}}                    & --       & 12B            & --                     & 0.16        &-3.83              & -0.87           & 62.26         &- \\
                            & \multicolumn{2}{c}{FLAN-T5 \cite{chung2022scaling}}                     & --       & 11B            & --                     & 0.13         &-4.57             & -1.23         & 64.92          &- \\
                              & \multicolumn{2}{c}{Koala \cite{koala_blogpost_2023}}                       & --       & 7B             & --                      & 0.19     &-3.85                 & -0.84                 & 63.96 &-  \\
                              & \multicolumn{2}{c}{Mosaic MPT \cite{mosaicml2023introducing}}                  & --       & 7B             & --                      & 0.14   &-3.72                   & -0.82               & 63.21    &-  \\
                              & \multicolumn{2}{c}{StableLM \cite{stablelm}}                    & --       & 7B             & --                      & 0.17       &-4.12               & -0.98       & 62.47            &- \\ 
                              & \multicolumn{2}{c}{ChatGLM \cite{du2022glm}}                     & --       & 6B             & --                      & 0.27      &-3.52                & -0.62      & 70.38             &- \\\hline
\multicolumn{2}{c}{\multirow{7}{*}{Ensemble}} & Classifier-OPT       & 125M     & \textbf{13B}    & \textbf{3.98}     &0.27  &-3.44    & -0.61  & 69.60  &2.84  \\
\multicolumn{2}{c}{}        & PairRanker \cite{jiang2023llm}             & 400M     & 117B    & 44.41   &0.32  &-3.14   & -0.38  & 73.03 &2.25  \\
\multicolumn{2}{c}{}        & LLM-debate \cite{du2023improving}     & -     & 234B    & 56.58   &0.27  &-3.51   & -0.55  & 71.59 &3.16 \\
\multicolumn{2}{c}{}        & ReConcile \cite{chen2023reconcile}     & -     & 351B    & 55.75   &0.24  &-3.79  & -0.71  & 68.61 &3.68 \\
\multicolumn{2}{c}{}        & sampling-voting \cite{li2024more}     & 110M     & 234B    & 38.07   &0.27 &-3.39   & -0.33  & 70.12 &3.20 \\
\multicolumn{2}{c}{}        & DyLAN \cite{liu2023dynamic}     & -     & \textgreater 234B    & 115.68   &0.28  &-3.69  & -0.64  & 70.90 &3.85\\
\rowcolor{pink!30}
\multicolumn{2}{c}{}       & \textbf{DER (Ours)}   & \textbf{125M}     & 17B    &9.32      &\textbf{0.35{\tiny (+9.38\%)}}    &\textbf{-3.14{\tiny (+0.00\%)}}  &\textbf{-0.31{\tiny (+6.06\%)}}     &\textbf{75.00{\tiny (+2.70\%)}}   &\textbf{2.02}  \\ 
\bottomrule[1.0pt]
\end{tabular}
\end{small}
\end{center}
\caption{Comparison with LLMs and state-of-the-art baselines on the \textit{MixInstruct}. $(\cdot)$ indicates relative improvement over the second-best.}
\label{table: compare}
\end{table*}

\subsection{Knowledge Transfer Prompt}
\label{subsec: ktp}
We develop the Knowledge Transfer Prompt (KTP) to facilitate effective knowledge transfer among LLMs. The proposed KTP expects that the current LLM effectively uses the answer (knowledge) of the previous LLM $\hat{y}_{t-1}$ without being limited by it, to improve the performance of generating a better answer to the input $x$. To ensure that LLM experts follow the knowledge transfer settings, we introduce a role-playing mechanism \cite{kong2023better} into the KTP as:

\begin{mybox}[title={Knowledge Transfer Prompt:}, coltitle=white, colbacktitle=black]
\textit{[$x$] \textbackslash n There is an answer to the question from another student: \textbackslash n [$\hat{y}_{t-1}$] \textbackslash n Using another student's answer as additional advice, you need to give a more satisfactory answer directly. DO NOT mention other students.}
\end{mybox}

First, we treat the previous LLM's answer $\hat{y}_{t-1}$ as the ``student's answer", thereby avoiding the overriding influence of the answer content of the predecessor. We then ask the current LLM to refer to the ``student's answer" to give a more satisfactory answer $\hat{y}_{t}$ to question $x$ via ``\textit{you need to give a more satisfactory answer}". In addition, the proposed KTP avoids LLM outputting role-playing messages and irrelevant information by ``\textit{DO NOT mention other students}".

\subsection{Learning with Proximal Policy Optimization}
We use the Proximal Policy Optimization (PPO) \cite{ppo} to optimize the parameters $\theta$ of the DER-Agent (policy) due to the stability and sample-efficiency as:

\textbf{Actor.} The actor (DER-Agent) is trained for LLM selection according to the question-answer pair. To enhance the ability of natural language understanding, we employ a pre-trained OPT-125M \cite{zhang2022opt} with two Linear layers connected to the last hidden layer, where the output dim is the number of Candidate LLMs $N$.
In the off-policy algorithm, the old policy $\pi_{\theta_{old}}$ with old parameters $\theta_{old}$ is used to collect trajectories with the environment, while the policy $\pi_{\theta}$ is updated using trajectories collected by $\pi_{\theta_{old}}$.

\textbf{Critic.} The critic $V_\phi(s)$ is used to estimate the expected return $v_t$ of the state $s_t$ and calculate the advantage, which aids the actor in learning more efficiently and stably. The critic is composed of an OPT-125M with two Linear layers. But the output dim is set to one. Besides, the old critic $V_{\phi_{old}}(s)$ is used to collect trajectories, and the new critic $V_{\phi}(s)$ is updated using the collected trajectories.

\textbf{Learning Objectives.}
The goal of the learning is to maximize the expected long-term
return $\mathcal{J}(\theta)$:
\begin{align}
    \mathcal{J}(\theta)&=\mathbb{E}_{\tau \sim \pi_\theta(\tau)}[G(\tau)] \nonumber\\
    &=\mathbb{E}_{\tau \sim \pi_{\theta_{old}}(\tau)}    [\min (\rho A^{\pi_{\theta_{old}}}(s_t, a_t), \nonumber\\ &\operatorname{clip}\left(\rho, 1-\epsilon, 1+\epsilon\right) A^{\pi_{\theta_{old}}}(s_t, a_t))],
\label{eq:Jtheta}
\end{align}
where $G(\tau)$ is the total return of the trajectory $\tau=\{s_t,a_t,r_t,v_t,A^{\pi_{\theta_{old}}}(s_t, a_t)\}$ obtained by $\pi_{\theta_{old}}$ and $V_{\phi_{old}}$, 
$\rho=\frac{\pi_\theta(a_t \mid s_t)}{\pi_{\theta_{old}}(a_t \mid s_t)}$ is the ratio of the probability of action $a_t$ given by  $\pi_\theta$ and $\pi_{\theta_{old}}$ for state $s_t$, and $\epsilon$ is a hyperparameter, usually set to 0.2 \cite{ppo}. 
The operation $\operatorname{clip}\left(\rho, 1-\epsilon, 1+\epsilon\right)$ constrains $\rho$ to the range $\left[1-\epsilon, 1+\epsilon\right]$, and
$A^{\pi_{\theta_{old}}}(s_t, a_t)=r_t - V_{\phi_{old}}(s_t)$ is the advantage at $t$.

\begin{algorithm}[!t]
    \caption{PPO Training for DER}
    \label{alg:training_algorithm}
    \begin{algorithmic}[1]
    \REQUIRE
    Prompt-Answer dataset $\mathcal{D}$, DER-Agent $\pi_{\boldsymbol{\theta}}$, critic $V_\phi$, replay buffer $\bm{\mathcal{B}}$, buffer size $M$, training iterations $m$, actor $\theta$  and critic $\phi$.
    
    \STATE Initialize $\bm{\mathcal{B}}$, parameters $\theta$  and $\phi$.
    
    \WHILE{Not converged}
        \FOR {$(question~x_i, answer~y_i)$ in $\mathcal{D}$}
            \STATE Collect a trajectory $\tau$ using old $\pi_{\theta_{old}}$ and $V_{\phi_{old}}$, and put it into $\bm{\mathcal{B}}$.
            
            \IF {$|\bm{\mathcal{B}}| = M$}
                \FOR{$iteration=1, 2, \dots, M$}
                
                    \STATE Uniformly  sample $\tau \in \bm{\mathcal{B}}$.
                    
                    \STATE Calculate $\mathcal{J}(\theta)$ via Eqn. (\ref{eq:Jtheta}).

                    \STATE Update $\theta$ to maximize $\mathcal{J}(\theta)$.
                    
                    \STATE Calculate TD error $\delta_t$ via Eqn. (\ref{eq:TD}).

                     \STATE Update $\phi$ to minimize TD error $\delta_t$.

                \ENDFOR
                \STATE Empty the replay buffer $\bm{\mathcal{B}}$.
                \STATE Update $\theta_{old} \leftarrow \theta$.
                \STATE Update $\phi_{old} \leftarrow \phi$.
            \ENDIF
        \ENDFOR
    \ENDWHILE
    \end{algorithmic}
\end{algorithm}

\begin{table*}[t]
\begin{center}
\renewcommand{\arraystretch}{0.98}
\renewcommand{\tabcolsep}{0.2pt}
\begin{small}
\begin{tabular}{ccccccccccc}
\toprule[1pt]
\multirow{2}{*}{Category}     & \multicolumn{2}{c}{\multirow{2}{*}{Methods}}    & \multicolumn{2}{c}{Param. $\downarrow$} & \multirow{2}{*}{Acc (\%) $\uparrow$} & \multirow{2}{*}{Rouge-S $\uparrow$} & \multirow{2}{*}{BARTScore $\uparrow$} & \multirow{2}{*}{BLEURT $\uparrow$}  & \multirow{2}{*}{BERTScore $\uparrow$} &\multirow{2}{*}{GPT-Rank$\downarrow$}\\ \cline{4-5}
                              & \multicolumn{2}{c}{}                            & Agent    & Infer.      &       &                     &                            & &  &                    \\ \hline
\multirow{11}{*}{LLMs}        & \multicolumn{2}{c}{MOSS \cite{sun2023moss}}                        & --       & 16B            & 19.41       &   0.29            & -5.27                      & -0.58      & 70.14  &-            \\
                              & \multicolumn{2}{c}{Vicuna \cite{chiang2023vicuna}}                      & --       & 13B        & 33.74     &  0.36               & -4.51                      & -0.43           & 74.68  &-       \\
                              & \multicolumn{2}{c}{Alpaca \cite{taori2023stanford}}                      & --       & 13B            & 8.11     &  0.17              & -5.86                      & -0.95             & 63.07 &-     \\
                              & \multicolumn{2}{c}{Baize \cite{xu2023baize}}                       & --       & 13B            &31.08       &  0.30             & -5.11                      & -0.59               & 69.97 &-   \\
                            & \multicolumn{2}{c}{Open Assistant \cite{openassistant}}              & --       & 12B            & 16.60      &      0.31          & -5.26                      & -0.58                 & 70.64 &-    \\
                            & \multicolumn{2}{c}{Dolly2 \cite{conover2023free}}                    & --       & 12B            & 11.52      &    0.23           & -5.66                      & -0.88                & 67.27 &-     \\
                            & \multicolumn{2}{c}{FLAN-T5 \cite{chung2022scaling}}                     & --       & 11B            & 24.03     &   0.37              & -4.94                      & -0.42             & 74.68     &-    \\
                              & \multicolumn{2}{c}{Koala \cite{koala_blogpost_2023}}                       & --       & 7B             & 20.70       & 0.25              & -4.94                      & -0.95          & 67.98    &-      \\
                              & \multicolumn{2}{c}{Mosaic MPT \cite{mosaicml2023introducing}}                  & --       & 7B             &23.43   & 0.31                  & -4.97                      & -0.53      & 70.77     &-         \\
                              & \multicolumn{2}{c}{StableLM \cite{stablelm}}                    & --       & 7B             & 19.56      &    0.32            & -4.70                     & -0.39                & 74.76 &-    \\ 
                              & \multicolumn{2}{c}{ChatGLM \cite{du2022glm}}                     & --       & 6B             & 21.08     &   0.33              & -4.90                      & -0.57               & 72.64 &-    \\\hline
\multicolumn{2}{c}{\multirow{5}{*}{Ensemble}}        & PairRanker \cite{jiang2023llm}      & 400M     & 117B    & 30.17 & 0.36 &-4.74     & \textbf{-0.39} & 74.26 &1.85  \\
\multicolumn{2}{c}{}        & LLM-debate \cite{du2023improving}     & -     & 234B    & 19.71 &0.27 &-5.35     & -0.74 & 68.39  &2.43  \\
\multicolumn{2}{c}{}        & sampling-voting \cite{li2024more}     & 110M     & 234B    & 26.08  & 0.36&-4.91     & -0.34 & 73.21  &1.68  \\
\multicolumn{2}{c}{}        & DyLAN \cite{liu2023dynamic}     & -     & \textgreater 234B    & 19.64 &0.25  &-5.41     & -0.77 & 68.02  &2.77  \\
\rowcolor{pink!30}
\multicolumn{2}{c}{}       & \textbf{DER (Ours)}   & \textbf{125M}     & \textbf{26B}    &\textbf{34.98{\tiny (+15.94\%)}} &\textbf{0.37} &\textbf{-4.44}   &-0.41  &\textbf{75.14} &\textbf{1.27} \\ 
\bottomrule[1.0pt]
\end{tabular}
\end{small}
\end{center}
\caption{Comparison with LLMs and state-of-the-art baselines on the \textit{GSM8K}.}
\label{Table: compare_gsm8k}
\end{table*}

\textbf{Training.} The overview of the optimization process is presented in Algorithm \ref{alg:training_algorithm}. Specifically, given a Prompt-Answer dataset $\mathcal{D}$, we use $\pi_{\theta_{old}}$ and $V_{\phi_{old}}$ to interact with the environment to collect a trajectory $\tau$ and compute the advantage $A^{\pi_{\theta_{old}}}(s_t, a_t)$. We then put $\tau$ into the reply buffer $\bm{\mathcal{B}}$. When a certain number of trajectories (such as $M$) have been collected, they are used to train the actor and critic. In particular, we first uniformly sample sequences from the reply buffer $\bm{\mathcal{B}}$, and then calculate the expected long-term return $\mathcal{J}(\theta)$ so that to optimize the parameters of the policy $\pi_{\theta}$. Besides, the Temporal Difference (TD) error $\delta_t$ is also calculated to optimize the parameters of the critic $V_{\phi}$:
\begin{equation}
    \delta_t=G_t-V_\phi\left(s_t\right),
    \label{eq:TD}
\end{equation}
where $G_t$ is the total expected return starting from time step $t$.
After training a certain number of times using the samples in the existing reply buffer $\bm{\mathcal{B}}$, we clear the reply buffer and update the parameters of the old policy $\pi_{\theta_{old}}$ and critic $V_{\phi_{old}}$. Then we repeat the above operation until convergence.

\section{Experiment}
\label{sec: experiment}

\textbf{Datasets and LLM experts.} 
Following the settings of PairRanker \cite{jiang2023llm}, we use \textit{MixInstruct} as the benchmark. In addition, we use the \textit{GSM8K} \cite{cobbe2021training} and \textit{Multidomain} we constructed (see Appendix 2.3) for further evaluation.
We select eleven LLM experts for the ensemble task: Open Assistant \cite{openassistant}, Vicuna \cite{chiang2023vicuna}, Alpaca \cite{taori2023stanford},  Baize \cite{xu2023baize}, MOSS \cite{sun2023moss}, ChatGLM \cite{du2022glm}, Koala \cite{koala_blogpost_2023}, Dolly2 \cite{conover2023free}, Mosaic MPT \cite{mosaicml2023introducing}, StableLM \cite{stablelm} and FLAN-T5 \cite{chung2022scaling}.

\textbf{Baseline Methods.} 1) Classifier-OPT (Appendix 2.5), 2) PairRanker \cite{jiang2023llm}, 3) LLM-debate \cite{du2023improving}, 4) ReConcile \cite{chen2023reconcile}, 5) sampling-voting \cite{li2024more}, and 6) DyLAN \cite{liu2023dynamic}.

\begin{table}[t]
\renewcommand{\arraystretch}{1.0}
\renewcommand{\tabcolsep}{3pt}
\begin{center}
\begin{small}
\begin{tabular}{cccc}
\toprule[1pt]
Version    & BARTScore $\uparrow$ & BLEURT $\uparrow$  & BERTScore $\uparrow$ \\ \hline
Random (w/o KTP)        &-3.48         &-0.45  &72.69  \\
Random      &-3.42          &-0.45  &72.88   \\
Ours (w/o KTP)        &-3.29          &-0.40  &74.30  \\
\rowcolor{pink!30}\textbf{Ours}       &\textbf{-3.14}   &\textbf{-0.31}  & \textbf{75.00}   \\ \bottomrule[1.0pt]
\end{tabular}
\end{small}
\end{center}
\caption{Experimental results on the effect of KTP.}
\label{table: ablation}
\end{table}

\subsection{Comparison Experiments}
\label{subsec: compare}
We compare our proposed DER, eleven LLM experts, ChatGPT, and state-of-the-art (SOTA) LLM ensemble methods to demonstrate our method superior performance. We conduct experiments on a variety of downstream tasks, including QA task (see Table \ref{table: compare}), mathematical reasoning task (see Table \ref{Table: compare_gsm8k}), and multi-domain QA task (see Appendix 3.2).

\textbf{DER is consistently better than single LLM.} From Table \ref{table: compare}, DER achieves better performance than a single LLM. Crucially, the DER on BARTScore is improved by 8.7\% compared to Vicuna. In addition, DER reaches 98\% of the ChatGPT performance on the BERTScore, while the inference parameters are only 10\% of those of ChatGPT. We conclude that DER achieves better performance than the single LLM due to its ability to aggregate the complementary knowledge of diverse LLMs through knowledge transfer.

\textbf{Trade-offs between performance and computational resources.}
As shown in Table \ref{table: compare}, the proposed DER achieves better performance than the SOTA methods with a little cost. Specifically, the proposed DER reduces the computational overhead of LLMs inference by about 85\% (117B $\rightarrow$ 17B) compared to PairRanker, while DER increases the BERTScore by about 2.7\% (73.03 $\rightarrow$ 75.00) compared to PairRanker. In addition, our method is also the best performer on the GPT-Rank metric. The main reason is that the reward function of our design requires DER-Agent to aggregate the strengths of diverse LLMs on as few resources as possible while setting the maximum route length. 

\textbf{Superior performance on mathematical reasoning.} From Table \ref{Table: compare_gsm8k}, DER outperforms SOTA methods on the mathematical reasoning task. Specifically, compared to PairRanker, DER reduces the inference cost by more than 77\% and increases the Accuracy by about 16\%. We conclude that DER effectively aggregates the strengths of LLMs to generate better mathematical reasoning results after the knowledge transmission through them.

\begin{table}[t]
\renewcommand{\arraystretch}{1.0}
\renewcommand{\tabcolsep}{6.0pt}
\begin{center}
\begin{small}
\begin{tabular}{cccccc}
\toprule[1pt]
$\alpha$ & $\beta $ & $\gamma$  & BARTScore $\uparrow$ & BLEURT $\uparrow$ &BERTScore $\uparrow$ \\ \hline
$\checkmark$   &$\checkmark$ &   &-3.32   &-0.34  &74.68  \\
$\checkmark$  & &$\checkmark$          &-3.22          &-0.34 & 74.52 \\
\rowcolor{pink!30} $\checkmark$   &$\checkmark$  & $\checkmark$    &\textbf{-3.14} &\textbf{-0.31}    &\textbf{75.00} \\ 
\bottomrule[1.0pt]
\end{tabular}
\end{small}
\end{center}
\caption{Experimental results for the component of the reward function on \textit{MixInstruct}. Notably, $\alpha=0.001$.}
\label{table: reward}
\end{table}

\subsection{Ablation Studies}
\label{subsec: ablation}
\textbf{Effectiveness of Knowledge Transfer Prompt.} 
We compare DER and DER (w/o KTP) to demonstrate the effectiveness of the KTP. From Table \ref{table: ablation}, the KTP effectively improves the performance of DER. Specifically, there is a 4.6\% increase in BARTScore after using the KTP. This is strong support for the fact that the use of a role-playing prompt allows the LLM to leverage the knowledge of the previous LLM to produce a more satisfactory output \cite{kong2023better}.

\textbf{Effectiveness of LLM ensemble using MDP.} 
As shown in Table \ref{table: ablation}, compared with the randomly generated route method, the DER outperforms the randomly generated route method by 2.31 in BERTScore. The primary reason is that by modeling the LLM ensemble as an MDP, the trained DER-Agent chooses an appropriate LLM based on the answer of the previous LLM to continue the answering.

\begin{table}[t]
\renewcommand{\arraystretch}{1.0}
\renewcommand{\tabcolsep}{0.2pt}
\begin{center}
\begin{small}
\begin{tabular}{ccccc}
\toprule[1pt]
Version & Param. $\downarrow$ & BARTScore $\uparrow$ & BLEURT $\uparrow$ & BERTScore $\uparrow$\\ \hline
$T_{max}=3$  &\textbf{15B}       & -3.15        &-0.32  &74.97   \\
\rowcolor{pink!30}$T_{max}=4$  &17B       &\textbf{-3.14}     &\textbf{-0.31} &\textbf{75.00}    \\
$T_{max}$ (w/o Term.)  &28B       &-3.15          &-0.34  &73.32   \\
$T_{max}=5$   &20B      &-3.16   &-0.32  & 74.94  \\ \bottomrule[1.0pt]
\end{tabular}
\end{small}
\end{center}
\caption{Experimental results with different reachable maximum step ($T_{\text{max}}$) and without Terminator (w/o Term.).}
\label{table: step}
\end{table}

\begin{table}[t]
\renewcommand{\arraystretch}{1.0}
\renewcommand{\tabcolsep}{10.0pt}
\begin{center}
\begin{small}
\begin{tabular}{ccccc}
\toprule[1pt]
Length   & $T=1$ & $T=2$ & $T=3$  &$T=4$\\ \hline
Percentage  &38.1\%       &17.6\%         &6.1\%   &38.2\% \\
\bottomrule[1.0pt]
\end{tabular}
\end{small}
\end{center}
\caption{Statistics of answer route length generated by DER on \textit{MixInstruct} testset for all samples.}
\label{table: area}
\end{table}

\begin{table}[t]
\renewcommand{\arraystretch}{1.0}
\renewcommand{\tabcolsep}{1.6pt}
\begin{center}
\begin{small}
\scalebox{0.98}{
\begin{tabular}{cccccc}
\toprule[1pt]
Version &Agent &Infer. & BARTScore $\uparrow$ & BLEURT $\uparrow$ &BERTScore $\uparrow$\\ \hline
Two experts   &125M  &49B   &-3.23 &-0.33   &74.80  \\
\rowcolor{pink!30} One expert   &125M  & \textbf{17B}    &\textbf{-3.14} &\textbf{-0.31}   &\textbf{75.00} \\ 
\bottomrule[1.0pt]
\end{tabular}
}
\end{small}
\end{center}
\caption{Experiments to the number of experts per state.}
\label{table: expert_num}
\end{table}

\textbf{Effects of the reward function $\bm{\mathcal{R}}$.} 
We study the effects of $\beta \Delta P({\hat y})$ and $\gamma$ in reward function on the performance of DER, by conditioning $\beta = 0$ or $\gamma = 0$ in Equation (\ref{equation: rt}) and (\ref{equation: rsa}). As shown in Table \ref{table: reward}, whenever $\beta = 0$ or $\gamma = 0$ both lead to about 3\% performance degradation of DER on the BARTScore metric. This provides strong support for adding additional rewards/ penalties to our reward function (see Section \ref{subsec: reward}) for answering the performance increase and whether or not completing the task within a finite step size improves DER's performance on the LLM ensemble task.

\textbf{Effects of the Terminator.} 
We study the effect of Terminator on DER at the maximum reachable step $T_{max}=4$. As shown in Table \ref{table: step}, DER with Terminator reduces the parameters by 39\% while also improving the BERTScore by 1.68. This is because the route reaches the endpoint when the Terminator believes that the output answer is greater than or equal to the threshold $p_0=0.73$ on the BERTScore metric (selection of $p_0$ see Appendix 4.2). Therefore, using Terminator reduces the average computational resources required for inference of LLMs and results in better performance.

\textbf{Effects of the maximum reachable step $T_{max}$.} 
As shown in Table \ref{table: step}, as the maximum reachable step increases, the average parameter of the LLMs inference also increases, but it does not make a significant difference to the performance. Therefore, we set the $T_{max}$ to 4. In addition, we count the length of answer routes generated by DER for all samples. As shown in Table \ref{table: area}, a substantial majority, i.e., 55.7\% of DER’s answer routes on \textit{MixInstruct} have a length $T\leq 2$, validating the practical efficiency of DER.

\textbf{Effects of the number of experts selected for each state.} We conduct experiments where two experts are selected in each state to demonstrate that selecting one expert per state is optimal. From Table \ref{table: expert_num}, DER produces better-generated answers by selecting an expert per state, with a 65\% reduction in inference parameters. 
The reason is that the knowledge gained from the non-optimal experts selected in each state affects the quality of the expert's answer in the next state.

\subsection{Eaxmple of DER}
\label{subsec: case}  
From Appendix 5, the DER-Agent selects a suitable LLM to continue the answering task as the sequence decision process proceeds, and the answering performance ends up being the best. Notably, the next LLM expert tends to leverage the knowledge of the previous LLM expert to improve the output answer. Thus, DER assembles complementary knowledge among LLMs to obtain better output answers.

\section{Conclusion}
\label{conclusion}
In this paper, we propose a novel dynamic ensemble reasoning method, called DER, to integrate the strengths of LLM experts dynamically conditioned on dynamic inputs. Specifically, we model the LLM ensemble as an MDP, where a DER-Agent takes dynamic inputs, sequentially asks an LLM candidate to provide knowledge, and passes the knowledge to subsequent LLM candidates. We devise a reward function to train a DER-Agent to select an optimal answering route given the input questions, aiming to achieve the highest performance with as little computational cost as possible. We develop a KTP that enables the subsequent LLM to utilize the expert knowledge of the prior LLMs. Experiments demonstrate that DER integrates the advantages of LLMs with only 15\% of the inference parameters compared to the LLM ensemble methods based on the output of all LLMs.

\section*{Acknowledgments}
This work was partially supported by the Joint Funds of the National Natural Science Foundation of China (Grant No.U24A20327), Key-Area Research and Development Program of Guangdong  Province (2018B010107001), and TCL Science and Technology Innovation Fund.

\section*{Contribution Statement}
This work was a collaborative effort by all contributing authors. Jinwu Hu, Yufeng Wang, and Shuhai Zhang made equal contributions to this study and are designated as co-first authors. Mingkui Tan and Bin Xiao, serving as the corresponding authors, are responsible for all communications related to this manuscript.

\bibliographystyle{named}
\bibliography{ijcai25}

\end{document}